\documentclass{article}

\usepackage[preprint]{neurips_2026}

\usepackage[utf8]{inputenc} 
\usepackage[most]{tcolorbox}
\usepackage[T1]{fontenc}    
\usepackage{multirow}
\usepackage{arydshln}
\usepackage{graphicx}
\usepackage{amsmath}
\usepackage{fancyvrb}
\usepackage{fvextra}
\usepackage{hyperref}       
\usepackage{tcolorbox}
\usepackage{url}            
\usepackage{subcaption}
\usepackage{booktabs}       
\usepackage{amsfonts}       
\usepackage{nicefrac}       
\usepackage{microtype}      
\usepackage{xcolor}         
\usepackage{amssymb}

\title{Verifiable Environments Are LEGO Bricks: Recursive Composition for Reasoning Generalization}

\vspace{5px}
\author{
\hspace{-2px}
Hao Xiang${}^{1,2}$\thanks{~ Equal contribution},
Qiaoyu Tang${}^{1,2}$\footnotemark[1],
Le Yu${}^{3}$,
Yaojie Lu${}^{2}$,
Xianpei Han${}^{2}$,
Ben He${}^{1,2}$
\\
\textbf{Le Sun}${}^{2}$,
\textbf{Bowen Yu}${}^{3}$,
\textbf{Peng Wang}${}^{3}$,
\textbf{Hongyu Lin}${}^{2}$,
\textbf{Dayiheng Liu}${}^{3}$ 
\vspace{5px}
\\
$^{\rm 1}$University of Chinese Academy of Sciences \vspace{3px}  \\
$^{\rm 2}$Chinese Information Processing Laboratory, Institute of Software, \\Chinese Academy of Sciences \vspace{3px}  \\
$^{\rm 3}$Qwen Team, Alibaba Group 
}

\begin{document}

\maketitle
\let\thefootnote\relax\footnotetext{Correspondence: \texttt{xianghao2022@iscas.ac.cn}}

\begin{abstract}
Reinforcement Learning (RL) with verifiable environments has emerged as a powerful approach for enhancing the reasoning capabilities of Large Language Models (LLMs).
While prior research demonstrates that scaling environment quantity improves RL performance, existing manual or individual construction methods suffer from linear scaling limits, thereby hindering scalable reasoning generalization. 
This paper introduces RACES (\textbf{R}ecursive \textbf{A}utomated \textbf{C}omposition for \textbf{E}nvironment \textbf{S}caling), a framework that conceptualizes verifiable environments as composable building blocks that can be recursively assembled.
The key insight is that when the codomain (output type) of one environment matches the domain (input type) of another, they can be automatically fused into a new verifiable environment, enabling recursive composition.
RACES is implemented with 300 individual environments and defines a set of composition operators (\textsc{SEQUENTIAL}, \textsc{PARALLEL}, \textsc{SORT}, and \textsc{SELECT}) that induce diverse reasoning patterns.
Extensive experiments show that RL training on these composite environments consistently enhances reasoning generalization. 
Specifically, RACES improves DeepSeek-R1-Distill-Qwen-14B by an average of 3.1 points (from 48.2 to 51.3) and boosts Qwen3-14B performance from 58.8 to 61.1 on six benchmarks, which are unseen during the construction of training environments. 
Moreover, RACES achieves performance comparable to training on 300 individual environments using only 50 base environments, demonstrating significant efficiency in environment utilization.
\end{abstract}

\section{Introduction}
Recent work has explored cost-effective methods for constructing reinforcement learning data with verifiable environments such as code, puzzles, and related tasks~\citep{zeng2025rlvescalingreinforcementlearning, wang2026rlanythingforgeenvironmentpolicy}, which yield deterministic outputs for given inputs. In this setting, models are required to reason over environment descriptions and inputs to produce correct outputs without relying on external tools~\citep{jiang-etal-2025-logicpro, ding2025longreasonarenalongreasoningbenchmark}.

Prior studies have demonstrated that increasing the number of environments leads to significant performance gains~\citep{zeng2025rlvescalingreinforcementlearning}, naturally motivating the scaling of environment pools. While several works have pursued limited scaling through automated environment synthesis~\citep{xu2026scalersyntheticscalableadaptivelearning,he2026resyn}, synthesizing environments individually increases the environment pool only linearly with construction cost. This linear scaling restricts the achievable diversity within a fixed budget and remains inadequate to support optimal reasoning generalization. Instead of generating entirely new environments from scratch, this study explores expanding the environment space by leveraging existing ones. Drawing inspiration from the closure property of transformation composition~\citep{birkhoff2017survey} and prior research on problem composition~\citep{yuan2026from,pei-etal-2025-mathfusion,chen-etal-2024-sifo}, this work introduces RACES (\textbf{R}ecursive \textbf{A}utomated \textbf{C}omposition for \textbf{E}nvironment \textbf{S}caling), a framework that composes existing environments to achieve scaling beyond linear growth.

The central insight of RACES is that when the codomain (output type) of one environment aligns with the domain (input type) of another, they can be fused into a new verifiable environment, enabling recursive composition.
As illustrated in Figure~\ref{fig:lego}, akin to LEGO bricks snapping together to form complex structures, verifiable environments can be assembled to create increasingly challenging reasoning tasks\footnote{If environments $f(x)$ and $g(x)$ each produce deterministic verifiable outputs, then their composition $g\circ f(x)=g(f(x))$ also yields a verifiable output via intermediate results.}.
To systematically convert these assembled composites into training problems, RACES defines a set of \emph{composition operators} including \textsc{SEQUENTIAL}, \textsc{PARALLEL}, \textsc{SORT} and \textsc{SELECT}, each specifying how an executable composite is presented to the model and how the response is verified.
For instance, \textsc{SEQUENTIAL} presents an assembled chain in execution order and requires the model to compute intermediate outputs, whereas \textsc{SORT} conceals the order and asks the model to recover a valid permutation that achieves a specified target output. By varying the operator choice and composition size, RACES scales to an unbounded space of environments and introduces diverse reasoning patterns, providing a robust basis for reasoning generalization.

To demonstrate the effectiveness of RACES, we construct a pool of 300 environments, implement four representative composition operators, and generate tens of thousands of composite environments for reinforcement learning on Qwen3~\citep{qwen3technicalreport} and DeepSeek-R1-Distill-Qwen~\citep{deepseekai2025deepseekr1incentivizingreasoningcapability}.
Across all model configurations, RACES consistently enhances generalization. For DeepSeek-R1-Distill-Qwen-14B, RACES increases the average score by 3.1 points (51.3 compared to 48.2), with notable improvements on IFEval (+4.0) and LongBench-v2 (+3.5). Similarly, RACES boosts Qwen3-14B performance (61.1 compared to 58.8).
Furthermore, RACES enables significantly more efficient environment utilization. Using only 50 base environments, it achieves performance comparable to training on 300 uncompounded environments.
Additional analysis demonstrates that the difficulty of composite environments can be systematically modulated by adjusting the composition size, offering greater flexibility for training design.
These findings establish RACES as an efficient and controllable approach for scaling verifiable environments and improving reasoning generalization.

\section{Related Work}
\paragraph{RLVR}
Reinforcement learning with verifiable rewards (RLVR) has become a leading approach for improving the reasoning capabilities of large language models~\citep{Guo2025, shao2024deepseekmathpushinglimitsmathematical, liu2025understanding}.
RLVR depends on a substantial volume of verifiable data to provide effective learning signals.
However, as model performance rapidly advances, even larger and harder curated datasets~\citep{he2026deepmathk} cannot keep up with the growing demand for high-quality training signals~\citep{villalobos2024position}.
Although several studies aim to improve training algorithms for more efficient use of existing data, these approaches do not directly address the issue of diminishing data effectiveness~\citep{zhu2025the, yue2025vapoefficientreliablereinforcement, wang2025reinforcement}.
Hence, automatic construction of verifiable training data has received considerable attention in recent years~\citep{zhao2025absolutezeroreinforcedselfplay, huang2026rzero, liang2026beyond}.

\paragraph{Verifiable Environments}
Generating training data via verifiable environments represents a promising direction in automatic data synthesis~\citep{jiang-etal-2025-logicpro}.
RLVE~\citep{zeng2025rlvescalingreinforcementlearning} introduces hundreds of algorithmically verifiable environments with adaptive difficulty, demonstrating the effectiveness of the verifiable environment approach by surpassing the performance of much larger static datasets.
SCALER~\citep{xu2026scalersyntheticscalableadaptivelearning} and RESYN~\citep{he2026resyn} further leverage LLMs to automatically generate reasoning environments equipped with verifiers, reducing the costs of manual environment construction.
Despite these advancements, current approaches scale the environment pool individually and fail to scale the environments towards infinite scale.
In contrast, RACES combines a finite set of environments to construct structurally diverse composites, inducing deeper and more varied reasoning patterns and scaling combinatorially with composition size.

\paragraph{Problem Composition}
Much research focuses on composing simple problems into more complex ones, thereby generating diverse training signals from limited data~\citep{pan2025reststresstestinglarge,zhou2025gsminfty,chen-etal-2024-sifo}.
MathFusion~\citep{pei-etal-2025-mathfusion} introduces strategies for pairwise composition of mathematical problems, emphasizing the critical role of composition method design.
H1~\citep{motwani2025h1bootstrappingllmsreason} composes GSM8K problems into extended dependency chains and applies curriculum reinforcement learning to enhance model capabilities.
Composition-RL~\citep{xu2026compositionrlcomposeverifiableprompts} sequentially integrates multiple verifiable problems to create more challenging tasks.
Although these methods show significant potential, they primarily focus on the question level, making it challenging to implement composition and resulting in shallow compositional depth. The full potential of composition remains unrealized~\citep{pei-etal-2025-mathfusion}.
Unlike these problem-level composition methods, RACES natively supports programmatic, recursive composition of verifiable environments, eliminating the need for hand-crafted adapters between each pair of composed objects.
\section{Method}
\subsection{Preliminary}
\label{sec:atomic}
The fundamental unit of RACES is the \textbf{verifiable environment}. Each verifiable environment $e$ is formally defined as a four-tuple:
\begin{equation}
e = \bigl(\, G_e, f_e, D_e, V_e\,\bigr),
\label{eq:atomic_env}
\end{equation}
where the components are defined as follows: 
\textbf{(1) Input sampler} $G_e$ programmatically yields valid input instances from $\mathcal{X}_e$, providing an unlimited stream of training data for each environment.
\textbf{(2) Output mapper} $f_e : \mathcal{X}_e \to \mathcal{Y}_e$ produces a unique output for any valid input. $f_e$ is implemented as code and encodes the core semantics of the verifiable environment. Its domain $\mathcal{X}_e$ and codomain $\mathcal{Y}_e$ jointly define the \textsc{domain signature} $\tau_e = (\mathcal{X}_e, \mathcal{Y}_e)$, which is the primary criterion for determining whether two environments can be composed.
\textbf{(3) Problem descriptor} $D_e : \mathcal{X}_e \to \Sigma^*$\footnote{$\Sigma^*$ denotes the natural language space.} utilizes a sampled input $x \in \mathcal{X}_e$ to render the environment to a solvable problem.
\textbf{(4) Programmatic verifier} $V_e: \mathcal{Y}_e \times \mathcal{Y}_e \to \{0, 1\}$ evaluates whether a model's output matches the reference answer $f_e(x)$.

This four-tuple characterizes a family of verifiable environments capable of generating an unlimited supply of RL training data and forms the foundation of RACES. Details regarding the construction of initial environments are provided in Appendix \ref{init_envs}.

\subsection{Compositional Closure of Verifiable Environments}
\label{sec:closure}
RACES leverages the compositional closure property of verifiable environments. Two environments $e_i$ and $e_j$ with domain signatures $\tau_{e_i}=(\mathcal{X}_{e_i}, \mathcal{Y}_{e_i})$ and $\tau_{e_j}=(\mathcal{X}_{e_j}, \mathcal{Y}_{e_j})$ are composable if the codomain of the first matches the domain of the second, that is, $\mathcal{Y}_{e_i}=\mathcal{X}_{e_j}$. Their composition is defined as:
\begin{equation}
(f_{e_j}\circ f_{e_i})(x)
=
f_{e_j}\bigl(f_{e_i}(x)\bigr),
\qquad x\in \mathcal{X}_{e_i}.
\label{eq:pairwise_composition}
\end{equation}
Because both mappers are deterministic, the composite $f_{e_j}\circ f_{e_i}$ is also deterministic, mapping $\mathcal{X}_{e_i}$ to $\mathcal{Y}_{e_j}$. Consequently, the composite maintains the same domain-codomain interface as the initial environment and can be further composed with additional environments.
More generally, a domain-compatible sequence $(e_1,\ldots,e_t)$ induces a composite mapper
\begin{equation}
F_{\pi_t}
=
f_{e_t}\circ f_{e_{t-1}}\circ \cdots \circ f_{e_1},
\label{eq:composite_mapping}
\end{equation}

with domain signature $(\mathcal{X}_{e_1},\mathcal{Y}_{e_t})$. If another environment $e'$ satisfies $\mathcal{X}_{e'}=\mathcal{Y}_{e_t}$, the composite can be recursively extended by $f_{e'}\circ F_{\pi_t}$.
This compositional closure enables RACES to expand a finite set of environments into a substantially larger space.
In practice, domain matching serves as a proposal mechanism rather than a guarantee of validity. The quality filtering process is detailed in Section \ref{sec:construction}.

\begin{figure}
    \centering
    \includegraphics[width=0.9\linewidth]{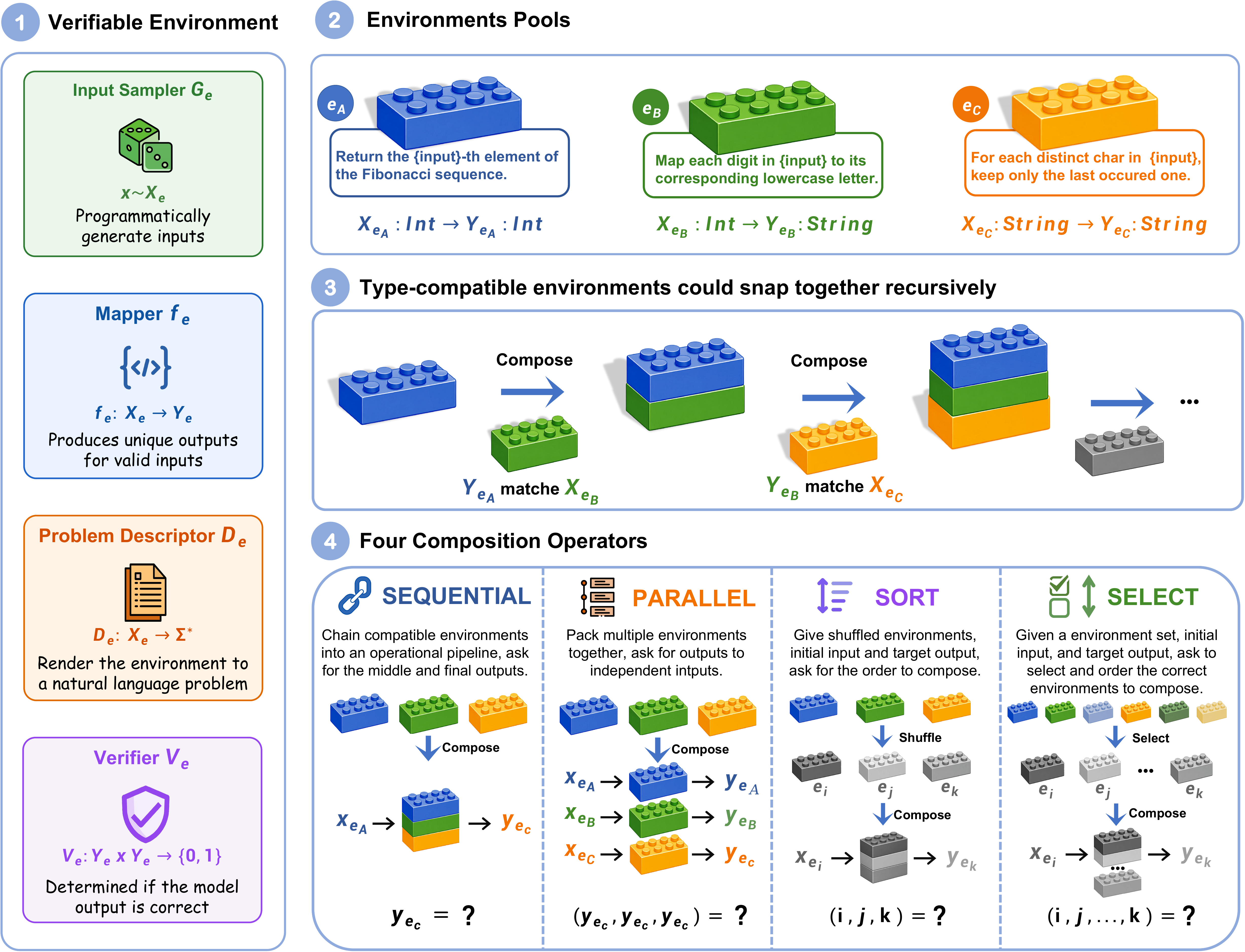}
    \caption{Overview of the RACES framework. 1) RACES standardizes the format of a verifiable environment, which consists of four components. 2) Following the standard interface, RACES constructs a pool of verifiable environments. 3) RACES can snap environments together like Lego bricks if the codomain of one matches the domain of another. 4) RACES implements four composition operators, resulting in diverse composite patterns and reasoning requests.}
    \label{fig:lego}
\end{figure}

\subsection{Implementation of Recursive Composition}
\label{sec:construction}

To operationalize compositional closure, RACES recursively constructs composite environments through a frontier-based search strategy.
Given an environment pool, RACES automatically assembles executable composite environments at scale and renders these as model-facing RL instances.
The construction process comprises three stages: composition path discovery, quality assurance, and operator instantiation.

\paragraph{Composition path discovery.}
Given an initial input $x_0$ and an environment pool $\mathcal{E}$, RACES searches over domain-compatible composition paths. All valid continuations form a search tree rooted at $x_0$: each node is a state value and each outgoing edge an environment whose domain matches that state. A length-$t$ path is represented as
\begin{equation}
\pi_t = (x_0;\ e_1,\ldots,e_t;\ y_1,\ldots,y_t),
\qquad y_i=f_{e_i}(\mathcal{Y}_{i-1}),\quad y_0=x_0,
\label{eq:path}
\end{equation}
yielding the composite mapper $F_{\pi_t}$ defined in Eq.~\eqref{eq:composite_mapping}. At each frontier state $y_t$, RACES retrieves all domain-compatible candidates $\mathcal{C}(y_t)=\{e\in\mathcal{E}:X_e=\operatorname{type}(y_t)\}$ and executes each on $y_t$. If execution succeeds and the output passes the later quality checks, the extended path
\begin{equation}
\pi_{t+1}=\mathrm{extend}(\pi_t, e)=(x_0;\ e_1,\ldots,e_t,e;\ y_1,\ldots,y_t,f_e(y_t))
\label{eq:path_extension}
\end{equation}
is added to the frontier for further extension.
RACES implements the search as a randomized breadth-first traversal, constrained by maximum composition depth, per-execution time limits, and a cap on extensions per frontier state. Each environment is assigned a usage budget, and candidates are sampled with probabilities weighted by remaining budgets to promote balanced utilization of the pool.

\paragraph{Quality assurance.}
Domain compatibility is a necessary but insufficient condition for effective composition.
Because each $f_e$ is implemented as executable code, a domain-compatible extension may still fail on specific intermediate states due to runtime exceptions, step-limit violations, timeouts, or invalid outputs.
RACES therefore applies online executable filtering during composition path discovery, retaining only those extensions that execute successfully and yield well-formed intermediate states.
This step ensures that the constructed composites are both domain-compatible and executable, making them suitable for programmatic verification. The complete criteria are provided in Appendix~\ref{app:quality}.

\paragraph{Operator instantiation.}
After trajectory discovery and quality assurance, RACES obtains a pool of executable composite environments that are not yet formatted as model-facing training problems.
A composition operator transforms a composite environment into a model-facing problem by specifying the presented information, required model predictions, and the verification process. A single composite environment can support multiple operators, resulting in distinct reasoning patterns without redefining the composition. The composition size $t$ directly measures task difficulty, as longer paths require the model to track additional intermediate states and mitigate error accumulation. RACES samples operator-specific sizes and applies balanced sampling to prevent the training set from being dominated by a narrow range of path lengths.

\subsection{Composition Operators}
\label{sec:operators}

As shown in Figure \ref{fig:lego}, RACES implements four representative operators: SEQUENTIAL, PARALLEL, SORT, and SELECT.

\paragraph{SEQUENTIAL}
Given $x_0$ and the ordered descriptors $(D_{e_1},\ldots,D_{e_t})$, the model predicts all intermediate outputs $(\hat y_1,\ldots,\hat y_t)$. The reward rewards the longest correct prefix:
\begin{equation}
K = \min\left(\{i\in\{1,\ldots,t\}:V_{e_i}(\hat y_i,y_i)=0\}\cup\{t+1\}\right)-1, \qquad R^{\mathrm{Seq}}=\frac{K}{t},
\label{eq:seq_reward}
\end{equation}
This reward structure captures the causal dependency in chained execution, where an error at any step invalidates all subsequent outputs.

\paragraph{PARALLEL}
PARALLEL evaluates the model's ability to maintain multiple independent computational threads within a shared context.
This operator samples $n$ independent environment-input pairs $(e_i,x_i)$ and asks the model to solve them jointly within a single context. The reward is computed as the average of individual verifications:
\begin{equation}
R^{\mathrm{Par}} = \frac{1}{n}\sum_{i=1}^{n}V_{e_i}(\hat y_i,y_i).
\label{eq:parallel_reward}
\end{equation}

\paragraph{SORT}
The model receives $x_0$, the terminal output $y_t$, and a shuffled set of environment descriptors, and must output a permutation $\hat{\sigma}$ of $\{1,\ldots,t\}$. The verifier executes the predicted order from $x_0$, and the reward is
\begin{equation}
R^{\mathrm{Sort}} = 
\begin{cases}
1, & \text{if } \bigl(f_{e_{\hat{\sigma}(t)}} \circ \cdots \circ f_{e_{\hat{\sigma}(1)}}\bigr)(x_0) = y_t, \\
0, & \text{otherwise.}
\end{cases}
\label{eq:sort_reward}
\end{equation}
Non-canonical but correct orderings also receive credit under this reward scheme.

\paragraph{SELECT}
SELECT requires the model to identify both the correct subset and the appropriate ordering. The path is augmented with distractors sampled from domain-compatible alternatives encountered during the search process:
\begin{equation}
\mathcal{A}_{\pi} = \{e_1,\ldots,e_t\} \cup \mathcal{B}_{\pi},
\label{eq:select_candidates}
\end{equation}
where $\mathcal{B}_{\pi}$ denotes the distractor set. Because the distractors are themselves executable on intermediate states, they are harder than random negatives. Given $x_0$, $y_t$, and $\mathcal{A}_{\pi}$, the model outputs an ordered sequence of distinct candidates $\hat{\sigma}=(\hat{\sigma}_1,\ldots,\hat{\sigma}_k)$ with each $\hat{\sigma}_i\in\mathcal{A}_{\pi}$. The reward is
\begin{equation}
R^{\mathrm{Sel}} = 
\begin{cases}
1, & \text{if } \bigl(f_{\hat{\sigma}_k} \circ \cdots \circ f_{\hat{\sigma}_1}\bigr)(x_0) = y_t, \\
0, & \text{otherwise.}
\end{cases}
\label{eq:select_reward}
\end{equation}
\section{Experiments}
\label{sec:experiments}
\subsection{Experimental Setup}
\label{sec:exp_setup}
\paragraph{Models.}
Main experiments are conducted on two 14B backbones: DeepSeek-R1-Distill-Qwen-14B and Qwen3-14B.
To reduce computational cost, analysis experiments are conducted on Qwen3-4B-Instruct-2507. 
All models are trained using the same pool of 300 verifiable environments described in Section \ref{sec:atomic}.

\paragraph{Evaluation}
Models are evaluated on six benchmarks: LiveCodeBench~\citep{jain2025livecodebench}, AIME 2024/2025~\citep{AIME2024I,AIME2025I}, Enigmata~\citep{chen2026enigmata}, IFEval~\citep{zhou2023instruction}, and LongBench-v2~\citep{bai-etal-2025-longbench}.
These benchmarks encompass code generation, mathematical reasoning, logic reasoning, instruction following, and long-context understanding.
None of these benchmarks are used in RACES. Consequently, improvements on these challenging benchmarks indicate genuine reasoning generalization.
To mitigate test randomness, each benchmark is evaluated multiple times (32 times for AIME 2024/2025 and 4 times for other benchmarks), and the average score is reported.

\paragraph{Data Construction.}
Two configurations sharing the same environment pool are compared: an individual-environment baseline that samples instances directly from the pool, and RACES, which samples a composition operator, a composition size, and domain-compatible environments to form a composite. Composition sizes are uniformly drawn from $[2,12]$ for \textsc{SEQUENTIAL} and \textsc{PARALLEL}, and $[2,6]$ for \textsc{SORT} and \textsc{SELECT}. In the analysis experiments, these ranges are reduced to $[2,6]$ for \textsc{SEQUENTIAL} and \textsc{PARALLEL}, and $[2,3]$ for \textsc{SORT}. An anonymized subset is available at \url{https://anonymous.4open.science/r/Submission_of_NIPS2026_34776-B7FD}.

\paragraph{Training Details.}
All RL experiments are implemented with VERL on a 32 × NVIDIA A100 80GB cluster and vLLM for rollout generation. Optimization is performed with GRPO~\citep{shao2024deepseekmathpushinglimitsmathematical} using a clip ratio of 0.28 and no KL regularization to a reference policy. For the main experiments, each configuration is trained for 300 steps using 12,800 training instances, a batch size of 128, a learning rate of 2e-6, 8 rollouts per problem, and a maximum sequence length of 32K tokens. For the analysis experiments, Qwen3-4B-Instruct-2507 is trained for 200 steps using 6,400 instances, a batch size of 64, and a maximum sequence length of 16K tokens. \textsc{Select} is excluded from the analysis as it yields near-zero rewards on this smaller backbone.

\subsection{Main Results}
\label{sec:main_results}
\begin{table}[!t]
\centering
\caption{\textbf{Main results.} LCBench and LBench-V2 denote LiveCodeBench and LongBench-V2; AIME is the average score on AIME 2024 and AIME 2025. Both RL settings use the same number of training instances and steps.}
\resizebox{0.8\linewidth}{!}{
\begin{tabular}{lcccccc}
\toprule
Model & LCBench & Enigmata & LBench-V2 & IFEval & AIME & Avg. \\
\midrule
\multicolumn{7}{l}
{\textcolor{gray}{\textit{DeepSeek-R1-Distill-Qwen-14B as base model}}}\\
\noalign{\vskip 0.11cm}
Base & 47.2 & 32.3 & 32.5 & 70.6 & 58.5 & 48.2 \\
$RL_{individual}$ & 46.9 & 34.2 & 33.7 & 69.3 & 59.8 & 48.8 \\
$RL_{RACES}$ & \textbf{48.8} & \textbf{35.4} & \textbf{36.0} & \textbf{74.6} & \textbf{61.7} & \textbf{51.3} \\
\noalign{\vskip 0.11cm}
\hdashline
\noalign{\vskip 0.11cm}
\multicolumn{7}{l}
{\textcolor{gray}{\textit{Qwen3-14B as base model}}}\\
\noalign{\vskip 0.11cm}
Base & 55.0 & 47.4 & 32.5 & 84.5 & 74.8 & 58.8 \\
$RL_{individual}$ & 56.3 & 48.2 & 34.1 & 85.7 & 76.0 & 60.1 \\
$RL_{RACES}$ & \textbf{57.0} & \textbf{49.2} & \textbf{35.5} & \textbf{86.7} & \textbf{77.0} & \textbf{61.1} \\
\bottomrule
\end{tabular}
}
\label{tab:main_results}
\end{table}

Table~\ref{tab:main_results} presents a comparison between RL on RACES ($RL_{RACES}$) and RL on individual environments ($RL_{individual}$).
$RL_{RACES}$ and $RL_{individual}$ are initialized from the same environment pool and utilize identical RL data scales.
Across both model families, RACES consistently demonstrates superior performance.
For DeepSeek-R1-Distill-Qwen-14B, $RL_{individual}$ increases the average score marginally, from 48.2 to 48.8.
In contrast, RACES improves the average score to 51.3, outperforming $RL_{individual}$ by 2.5 points.
For Qwen3-14B, RACES also surpasses $RL_{individual}$, increasing the average score from 60.1 to 61.1.
Performance gains are consistent across all benchmarks for both backbones.

$RL_{individual}$ exposes the model to isolated environments, whereas $RL_{RACES}$ trains the model on composed environments that require chaining transformations, maintaining intermediate states, and inferring valid operation orders or subsets.
The consistent advantage of RACES indicates that environmental composition offers a more effective training signal than sampling instances from the original environment pool, resulting in stronger transfer on unseen reasoning benchmarks.
\section{Analysis}
\label{sec:analysis}

The main experiments show that RACES improves generalization.
Further analysis is conducted to evaluate RACES's performance.
All analysis experiments are conducted on Qwen3-4B-Instruct-2507 to reduce computational cost.

\subsection{Performance across Training Process}
\label{sec:training_dynamic}

\begin{figure}[t]
\centering
\begin{subfigure}[t]{0.42\textwidth}
\centering
\includegraphics[width=\linewidth]{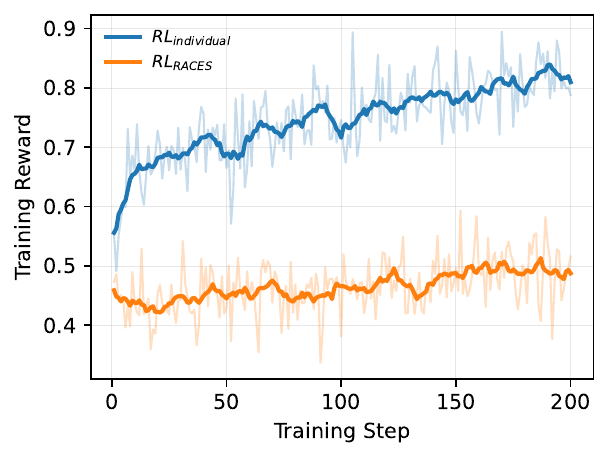}
\caption{Training Rewards.}
\label{fig:training_dynamic}
\end{subfigure}
\hspace{10px}
\begin{subfigure}[t]{0.42\textwidth}
\centering
\includegraphics[width=\linewidth]{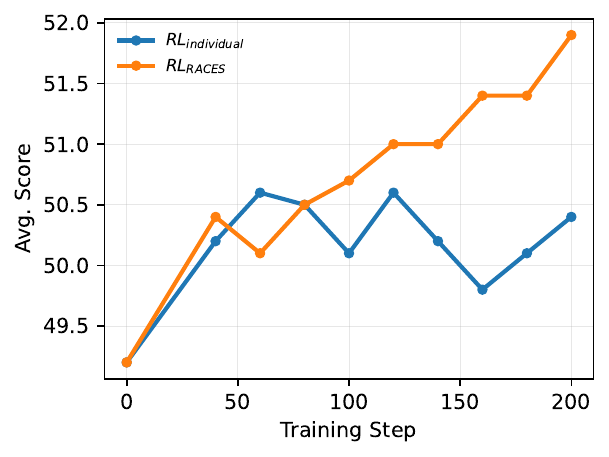}
\caption{Average performance across the training process.}
\label{fig:training_ood}

\end{subfigure}
\caption{\textbf{Training dynamics on Qwen3-4B-Instruct-2507.} \emph{(a)} Smoothed training reward over $200$ RL steps. \emph{(b)} Average performance.}
\vspace{-8pt}

\end{figure}

We first compare the training dynamics of $RL_{RACES}$ and $RL_{individual}$ on Qwen3-4B-Instruct-2507.
A natural expectation is that a method that increases reward more rapidly during reinforcement learning will also achieve superior downstream generalization.
However, the results indicate that this expectation does not hold. Figure~\ref{fig:training_dynamic} compares the training reward and average performance during RL training.
$RL_{individual}$ improves more rapidly and maintains consistently higher rewards throughout training, suggesting that the model adapts more easily to environments.
In contrast, $RL_{RACES}$ exhibits progressively slower reward growth, reflecting the increased complexity of composite environments.

However, the advantage of $RL_{individual}$ in training reward does not result in superior downstream performance.
As shown in Figure~\ref{fig:training_ood}, the average score of $RL_{individual}$ improves fast at the beginning, and then quickly peaks around 50.5.
Conversely, $RL_{RACES}$ continues to improve throughout the training process. While initially comparable to $RL_{individual}$, it gradually surpasses $RL_{individual}$ after the early stages, reaching 51.9 at step 200 compared to 50.4.

These findings differentiate between overfitting to training environments and acquiring transferable reasoning behaviors.
Individual environments are shorter and structurally simpler, making them easier to optimize and leading to rapid reward gains.
However, this improvement remains largely restricted to the training distribution.
Composite environments in RACES require the model to maintain intermediate states, execute multi-step transformations, and, in some cases, infer the latent order of operations.
Although these tasks are more challenging and result in slower reward improvement, they offer a richer training signal and produce more sustained gains on downstream benchmarks.
This suggests that the primary advantage of RACES is the induction of reasoning patterns that generalize more effectively beyond the training environments.

\subsection{Efficiency of Environment Utilization}
\label{sec:efficiency}

\begin{table}[!t]
\centering
\caption{Results of RACES and individual RL on the initial environment pool with different sizes.}
\resizebox{0.8\linewidth}{!}{
\begin{tabular}{lcccccc}
\toprule
Setting & LCB & Enigmata & LBench-V2 & IFEval & AIME & Avg. \\
\midrule
\multicolumn{7}{l}
{\textcolor{gray}{\textit{Qwen3-4B-Instruct-2507 as base model}}}\\
\noalign{\vskip 0.11cm}
\textsc{BASE} & 32.4 & 36.5 & 41.3 & 82.2 & 53.7 & 49.2 \\
$RL_{individual}$ (50 envs)  & 32.7 & 38.6 & 41.6 & 82.9 & 55.1 & 50.2  \\
$RL_{individual}$ (300 envs) & 31.5 & 39.3 & 41.9 & 83.5 & 55.6 & 50.4  \\
$RL_{RACES}$ (50 envs)       & 33.2 & 38.7 & 42.4 & 82.8 & 57.0 & 50.8  \\
$RL_{RACES}$ (300 envs)      & \textbf{34.0} & \textbf{39.5} & \textbf{42.7} & \textbf{85.2} & \textbf{58.3} & \textbf{51.9} \\
\noalign{\vskip 0.11cm}
\hdashline
\noalign{\vskip 0.11cm}
\multicolumn{7}{l}
{\textcolor{gray}{\textit{DeepSeek-R1-Distill-Qwen-14B as base model}}}\\
\noalign{\vskip 0.11cm}
\textsc{Base} & 47.2 & 32.3 & 32.5 & 70.6 & 58.5 & 48.2 \\
$RL_{individual}$ (300 envs) & 46.9 & 34.2 & 33.7 & 69.3 & 59.8 & 48.8 \\
$RL_{RACES}$ (50 envs)       & \textbf{47.4} & \textbf{34.7} & \textbf{35.7} & \textbf{73.0} & \textbf{60.3} & \textbf{50.2} \\
\bottomrule
\end{tabular}
}
\label{tab:env_efficiency}
\vspace{-10pt}
\end{table}

The results in Figure \ref{fig:training_ood} indicate that $RL_{RACES}$ significantly outperforms $RL_{individual}$ when both are provided with the same 300 initial environments.
Further analysis examines whether RACES maintains this advantage across varying numbers of environments.

Table~\ref{tab:env_efficiency} compares the performance of Qwen3-4B-Instruct-2507 on 50 and 300 initial environments, denoted as $RL_{setting}(50)$ and $RL_{setting}(300)$
$RL_{individual}$ improves the average score from 49.2 to 50.2 and rises to 50.4 when utilizing all 300 environments.
In contrast, $RL_{RACES}(50)$ achieves 50.8, outperforming $RL_{individual}(300)$ with only one-sixth of the base environment pool.
$RL_{RACES}(300)$ further increases to 51.9.
The same trend appears on DeepSeek-R1-Distill-Qwen-14B. $RL_{RACES}(50)$ reaches 50.2, clearly above $RL_{individual}(300)$ at 48.8.

This substantial difference demonstrates that RACES enhances the efficiency of environment utilization.
An individual environment can provide only instances of a single transformation family.
However, once composed, the same environment can appear at different positions, with different neighboring environments, under different operators.
Thus, composition converts a fixed environment pool into a much larger and structurally more diverse training distribution.
This represents the principal advantage of RACES over linear environment scaling, as it increases the utility of each existing environment rather than relying exclusively on synthesizing additional independent environments.

\subsection{Effect of Composition Size}
\label{sec:composition_size}

The effect of composition size on training effectiveness is subsequently examined.
Composition size is a central control variable in RACES, increasing it does not merely change the amount of generated data but directly changes the structure and difficulty of the resulting composite environments.
Larger compositions require the model to maintain more intermediate states, handle longer dependency chains, and tolerate stronger error propagation, thereby increasing the overall reasoning burden.

\begin{figure}
\centering
\begin{minipage}[t]{0.75\linewidth}
\vspace{0pt}
\centering
\includegraphics[width=\linewidth]{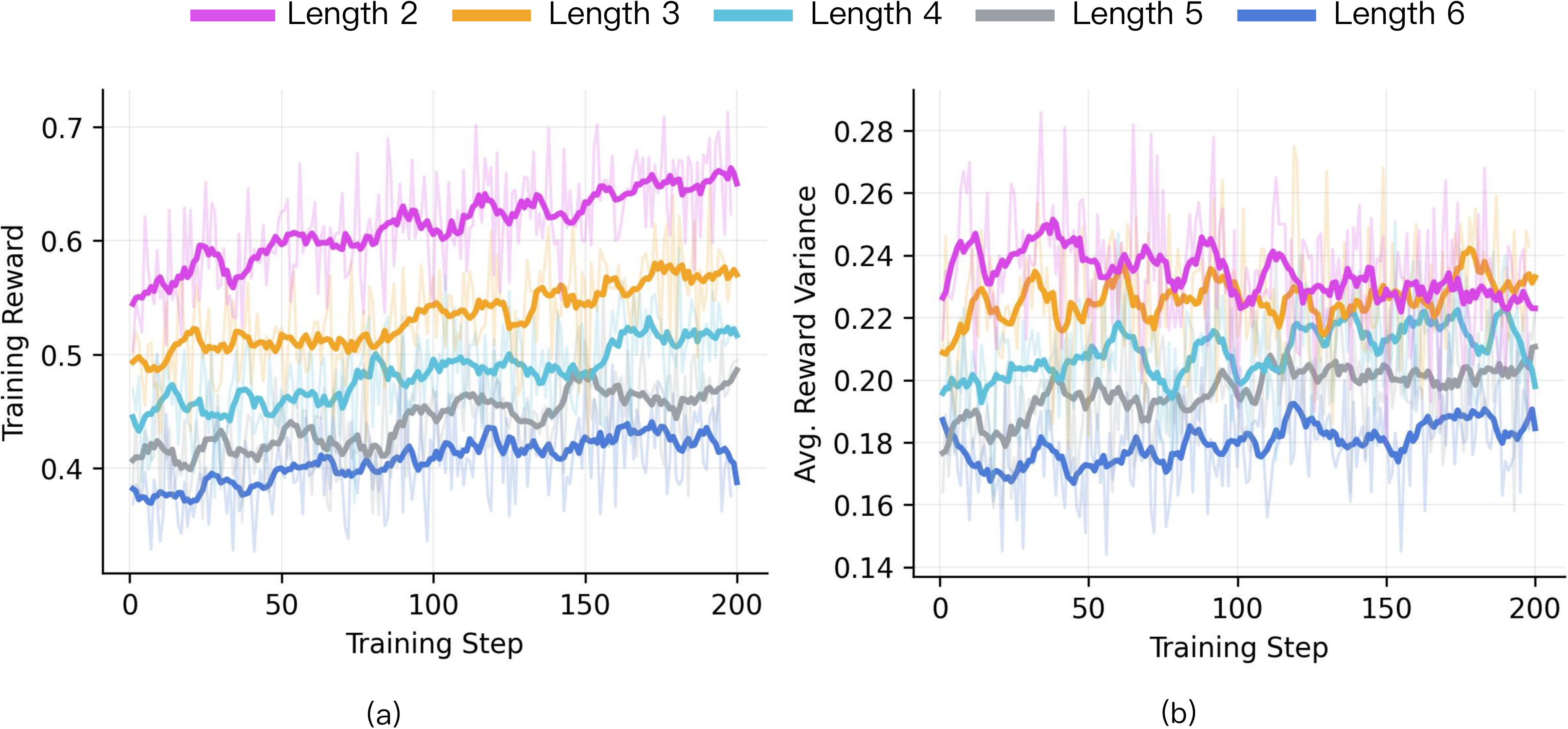}
\caption{Effect of composition size. \emph{(a)} rewards during training. \emph{(b)} Average reward variance, the mean of every problem’s rollout reward std.}
\label{fig:composition_size}
\end{minipage}\hfill
\begin{minipage}[t]{0.23\linewidth}
\vspace{0pt}
\centering
\captionof{table}{Average scores under different sizes.}
\label{tab:composition_size}
\small
\begin{tabular}{cc}
\toprule
Size & Performance \\
\midrule
2 & 50.8 \\
3 & 50.7 \\
4 & 51.0 \\
5 & \textbf{51.2} \\
6 & 50.7 \\
\bottomrule
\end{tabular}
\end{minipage}
\end{figure}

To isolate this effect, the environment pool is fixed at $|\mathcal{E}|=300$, and variants with different composition sizes are trained on Qwen3-4B-Instruct-2507.
The analysis focuses on the \textsc{SEQUENTIAL} operator, for which composition size has the most direct interpretation, and varies the size from 2 to 6.
Figure~\ref{fig:composition_size} summarizes both the training dynamics and the final performance under different composition sizes.

The training curves reveal a clear increase in optimization difficulty as composition size grows.
As shown in Figure~\ref{fig:composition_size}(a), the rewards are consistently higher for smaller composition sizes and decrease as the size increases, indicating that deeper compositions are harder for the policy to optimize.
A similar trend is observed in Figure~\ref{fig:composition_size}(b). The average reward variance (affecting advantage directly) tends to be lower for larger composition sizes, suggesting that deeper compositions leave a less informative learning signal for RL.

The effect on downstream performance is shown in Table~\ref{tab:composition_size}.
As the composition size increases from 2 to 5, the average score improves steadily from 50.8 to 51.2.
However, when the size is increased further to 6, performance decreases to 50.7.
This produces a clear non-monotonic trend. Increasing composition size is beneficial up to a moderate-to-large range, but excessively deep compositions become less effective.
Taken together, these results reveal a trade-off between reasoning generalization and trainability.
Small composition sizes are easier to optimize, but they provide only limited expansion beyond environments.
Increasing the composition size introduces richer multi-step reasoning patterns and improves reasoning generalization.
However, once the composition becomes too deep, the difficulty of optimization begins to dominate: rewards decrease, reward variance becomes less informative, and final performance deteriorates.
In practice, these results suggest that composition size should be regarded as a controllable curriculum variable rather than a parameter to maximize indiscriminately.


\subsection{Pattern Analysis}
\label{sec:pattern_analysis}

To complement the aggregate results, qualitative comparisons are made between responses from the base model, $RL_{individual}$, and $RL_{RACES}$ on identical AIME and Enigmata prompts. Three representative cases are presented in Appendix~\ref{app:cases}.

For AIME. In the $2{\times}2$ grid-coloring problem (Appendix~\ref{app:case-grid-coloring}), the base model sets up a valid enumeration but confuses variables when filling the case table and $RL_{individual}$ makes a single inconsistent update on an internal edge color. Both errors are local yet propagate to the final count. The $RL_{RACES}$ model instead introduces a compact function $f(x,y)$ that maps each pair of internal-edge colors to its boundary-assignment count, applies it uniformly, and cross-checks the total via an independent grouping calculation—replacing error-prone table-filling with representation-mediated counting and self-consistency verification.

For Enigmata. In a list-transformation task (Appendix~\ref{app:case-list-transform}), the base model and $RL_{individual}$ both cycle through surface rules (suffix extraction, duplicate-run selection) without rejecting hypotheses that fail earlier examples and $RL_{individual}$ even notes a contradiction yet continues with the falsified rule. The $RL_{RACES}$ model maintains explicit index–value tracking, tests each candidate against all demonstrations, and identifies the correct rule. A related pattern appears on Sum Skyscraper (Appendix~\ref{app:case-skyscraper}), non-RACES models commit to an incomplete row-wise search and reach a spurious “no valid solution.” $RL_{RACES}$ reframes the search column-by-column, exhausts a smaller space, propagates fixed values through row-uniqueness, and verifies all 16 clues.

These behaviors reflect the RACES training signal. \textsc{SEQUENTIAL} compositions train state carry-over, \textsc{Parallel} compositions train subproblem separation, \textsc{Sort} compositions train order inference, and \textsc{Select} introduces distractor discrimination. The transferable routines, including stable intermediate representations, constraint preservation, hypothesis revision, and final-answer verification, account for the observed performance gains without requiring surface similarity to the training environments.
\section{Conclusion}
\label{sec:conclusion}

We introduce RACES, a recursive composition framework that enables the modular assembly of verifiable environments. When the codomain of one environment aligns with the domain of another, the two can be composed into a new, verifiable environment that remains composable.
Our experiments demonstrate that RACES consistently enhances generalization across diverse model backbones and evaluation benchmarks.
Composing only 50 base environments with RACES outperforms direct reinforcement learning applied to 300 initial environments, highlighting the efficiency of RACES in leveraging available environments.
Further analysis indicates that the size of the composition serves as a principled control for task difficulty, with moderate composition depths resulting in optimal transfer performance.
Together, these results establish RACES as an efficient and controllable method for scaling verifiable environments beyond linear expansion.


\bibliographystyle{plainnat}
\bibliography{reference}

\newpage
\appendix
%
%
\section{Limitations}
\label{sec:limitations}

Several limitations of the present work are acknowledged.
First, the four composition operators address representative patterns but do not fully capture the entire design space.
More complex structures, such as conditional branching and bounded loops, remain unexplored.
Second, the effectiveness of RACES relies on the underlying model demonstrating adequate reasoning competence. Models with limited capabilities typically yield sparse rewards on composite tasks. Finally, composite environments generate extended reasoning traces, which require a sufficiently large context window (32K in these experiments) to enable effective training.
These limitations present opportunities for future research rather than imposing fundamental constraints on the framework.

\section{Declaration of LLM Usage}
\label{sec:llm_declaration}

\paragraph{Methodological use.} Following the SCALER~\citep{xu2026scalersyntheticscalableadaptivelearning} and RESYN~\citep{he2026resyn} synthesis paradigms, Claude-Sonnet-4.5 generates candidate verifiable environments, which subsequently pass through our two-stage filtering (static code self-consistency followed by multi-sample output consistency) before entering the initial environment pool.
\paragraph{Writing assistance.} LLMs were used exclusively for limited writing assistance, specifically for grammar checking and minor sentence-level refinement.

\section{Broader Impacts}
\label{sec:broader_impacts}

\paragraph{Positive impacts.} RACES advances automatic verifiable RL data construction, reducing reliance on human annotation. By transforming a small pool of environments into a structurally diverse and effectively unbounded space of composite tasks, RACES increases the accessibility of verifiable RL for research groups lacking industrial-scale annotation resources.

\paragraph{Potential negative impacts.} As with most foundational work on improving LLM capabilities, the techniques that enhance reasoning may also be misapplied. Specifically, the methodology could be adapted to synthesize verifiable training data for malicious purposes, and the energy consumption associated with large-scale RL training may contribute to environmental impact. Since model checkpoints are not released, the most direct vector of misuse is absent in this work. The remaining artifacts serve as general-purpose research infrastructure rather than capability-amplifying weights. The 300 environments do not contain content related to poison or dangerous material.

\section{Initial environments construction.}
\label{init_envs}
RACES constructs an initial pool $\mathcal{E}$ of 300 environments, sourced from three categories: (1) standardized algorithmic problem datasets~\citep{xia2025leetcodedatasettemporaldatasetrobust}\footnote{45 of the 300 environments are derived from algorithms.}; (2) environments automatically generated by Claude-Sonnet-4.5, following the synthesis paradigms of SCALER~\citep{xu2026scalersyntheticscalableadaptivelearning} and RESYN~\citep{he2026resyn}; and (3) manually authored environments.
Each environment undergoes a quality assurance review. For each, 20 inputs are sampled from $G_e$, and Claude-Sonnet-4.5 is tasked with solving $D_e(x)$ multiple times, requiring a $V_e$-judged pass rate exceeding 95
\label{app:quality}

This section outlines the filtering criteria applied to each candidate extension during composition-path discovery.

\paragraph{Executability.} Each candidate is executed on the current state within a sandboxed process and is discarded if it raises an exception, exceeds a 2-second wall-clock timeout, or returns no output. Extensions requiring more than 400 atomic steps are also rejected, as this is considered beyond the reasonable effort of a human solver.

\paragraph{Non-degeneracy.} Let $y_{t+1}$ denote the resulting output and $s = \mathrm{str}(y_{t+1})$ its string representation. An extension is discarded if $y_{t+1}$ (i) repeats an earlier state in the same path or duplicates a sibling output at the same frontier, (ii) equals $0$ or $1$, or is an integer greater than $500$, (iii) has a length exceeding $100$, or (iv) consists of a single repeated character.

After path discovery, paths with a length less than two are discarded, as single-step paths reduce to environments.

\section{Case Studies for Pattern Analysis}
\label{app:cases}

This appendix presents the prompts and key response excerpts for the three cases referenced in Section~\ref{sec:pattern_analysis}.
The excerpts focus on the decision points where the three models diverge; complete responses are released in the supplementary materials. We label the models as Base, $RL_{individual}$, and $RL_{RACES}$ for consistency with the main text.

\subsection{AIME: \texorpdfstring{$2\times2$}{2x2} Grid-Coloring}
\label{app:case-grid-coloring}

\textbf{Pattern illustrated.} A shift from ad-hoc enumeration toward representation-mediated counting: $RL_{RACES}$ introduces a reusable function abstraction that decouples argument tracking from arithmetic and verifies the final sum in two independent ways.

\smallskip
\begin{tcolorbox}[
  enhanced, breakable, width=\linewidth,
  colback=gray!5, colframe=gray!50,
  title={Problem (AIME 2025), ground truth: $\boxed{82}$},
  fonttitle=\bfseries\sffamily, coltitle=black
]
Four unit squares form a $2 \times 2$ grid. Each of the 12 unit line
segments forming the sides of the squares is colored either red or
blue in such a way that each unit square has 2 red sides and 2 blue
sides. Find the number of such colorings.
\end{tcolorbox}

\smallskip
\begin{tcolorbox}[
  enhanced, breakable, width=\linewidth,
  colback=red!3, colframe=red!40,
  title={\textbf{Base}, variable-confusion in case-table filling},
  fonttitle=\sffamily, coltitle=black
]
\begin{Verbatim}[fontsize=\footnotesize, breaklines, breakanywhere]
[Sets up grid, labels edges, identifies 4 internal + 8 boundary edges correctly.]

Each square must have exactly 2 red edges. [...] For each square, given the internal edges, the red edges in the square are determined by the pair. [...]

We can consider the possible red edge configurations for each square and see when the shared edges match. [...]

[Derives four equations:]
- (1): $x_1 + x_5 = 2 - a - c$, i.e. $s_1 = 2 - a - c$
- (2): $x_2 + x_7 = 2 - b - c$
- (3): $x_3 + x_6 = 2 - a - d$
- (4): $x_4 + x_8 = 2 - b - d$

[Builds a 16-row table of $(a,b,c,d)$ values:]

| a | b | c | d | s1 | s2 | s3 | s4 | w1 | w2 | w3 | w4 | f |
|---|---|---|---|----|----|----|----|----|----|----|----|---|
|0|0|0|0|2|2|2|2|1|1|1|1|1|
|0|0|0|1|2|2|1|1|1|1|2|2|4|
|**0**|**0**|**1**|**0**|**1**|**1**|**1**|**1**|**2**|**2**|**2**|**2**|**16**|
[...]

**<- KEY ERROR:** For $(a,b,c,d)=(0,0,1,0)$, RAW writes $s_3=2-a-d=2-0-0=1$ and $s_4=2-b-d=2-0-0=1$. The correct values are $s_3=2$ and $s_4=2$ (since $d=0$), giving $f=4$ not $16$. This variable-confusion error recurs in $\geq$8 of 16 rows.

[Sums all $f$ values to get 158.]

$$\boxed{158}$$
\end{Verbatim}
\end{tcolorbox}

\smallskip
\begin{tcolorbox}[
  enhanced, breakable, width=\linewidth,
  colback=orange!3, colframe=orange!50,
  title={\textbf{$RL_{individual}$}, single inconsistent state update},
  fonttitle=\sffamily, coltitle=black
]
\begin{Verbatim}[fontsize=\footnotesize, breaklines, breakanywhere]
[Uses case analysis by number of red internal edges $i_r \in \{0,1,2,3,4\}$.]

**Case $i_r=0$:** All internal blue. Each square forces both boundary edges red. -> **1 coloring** $\checkmark$

**Case $i_r=1$:** [Correctly derives constraints for H3 red:] $h1+v1=1$, $h2+v5=2$, $h5+v2=1$, $h6+v6=2$ -> $2 \times 2 = 4$ per choice, $\times 4$ choices -> **16** $\checkmark$

**Case $i_r=2$:**
- Opposite pairs: each gives $2^4=16$ -> $2 \times 16 = 32$ $\checkmark$
- Adjacent pairs: each gives $4$ -> $4 \times 4 = 16$ $\checkmark$
- Subtotal: **48** $\checkmark$

**Case $i_r=3$:** 4 choices $\times$ 4 each -> **16** $\checkmark$

**Case $i_r=4$:** All internal red.

[...]

- D: edges: H4 (red), H6, V4 (red), V6 -> red: $1 + h6 + **0** + v6$

**<- KEY ERROR:** Writes V4's contribution as $+0$ (treating it as blue), despite explicitly stating "V4 all red" at the start of this case. Correct: $1+h6+1+v6=2 \Rightarrow h6+v6=0 \Rightarrow$ **1 solution**, not $h6+v6=1 \Rightarrow$ 2 solutions.

[Sums: $1+16+48+16+2 = 83$]

$$\boxed{83}$$
\end{Verbatim}

\end{tcolorbox}

\smallskip
\begin{tcolorbox}[
  enhanced, breakable, width=\linewidth,
  colback=green!3, colframe=green!75!black,
  title={\textbf{RACES}, functional abstraction with two-way verification},
  fonttitle=\sffamily, coltitle=black
]
\begin{Verbatim}[fontsize=\footnotesize, breaklines, breakanywhere]

[Sets up same grid and equations, then introduces a key abstraction:]

For any two bits $x,y \in \{0,1\}$, define:
$$f(x,y) = \begin{cases} 1 & \text{if } x+y \in \{0,2\} \\ 2 & \text{if } x+y=1 \end{cases}$$

Then:
- $N_1 = f(a,c)$, $N_2 = f(b,c)$, $N_3 = f(a,d)$, $N_4 = f(b,d)$

Total = $\sum_{a,b,c,d} f(a,c) \cdot f(b,c) \cdot f(a,d) \cdot f(b,d)$

[Evaluates all 16 rows cleanly — each is a direct lookup of $f$:]

| a | b | c | d | f(a,c) | f(b,c) | f(a,d) | f(b,d) | Product |
|---|---|---|---|--------|--------|--------|--------|---------|
| 0 | 0 | 0 | 0 | 1 | 1 | 1 | 1 | 1 |
| 0 | 0 | 0 | 1 | 1 | 1 | 2 | 2 | 4 |
| 0 | 0 | 1 | 0 | 2 | 2 | 1 | 1 | **4** |
[... all 16 rows computed correctly ...]

[**Verification by grouping:**]
- 12 rows with product 4 -> $12 \times 4 = 48$
- 2 rows with product 16 -> $2 \times 16 = 32$
- 2 rows with product 1 -> $2 \times 1 = 2$
- Total: $48 + 32 + 2 = 82$ $\checkmark$

[**Sequential verification:**]
$1+4+4+16+4+4+4+4+4+4+4+4+16+4+4+1 = 82$ $\checkmark$

$$\boxed{82}$$

\end{Verbatim}
\end{tcolorbox}

\subsection{Enigmata: List-Transformation Rule Induction}
\label{app:case-list-transform}

\textbf{Pattern illustrated.} Hypothesis verification against all
demonstrations, combined with explicit index--value tracking.
$RL_{RACES}$ systematically rules out candidate rules that fail on
earlier examples, where Base and $RL_{individual}$ repeatedly
propose rules contradicted by their own observations.

\smallskip
\begin{tcolorbox}[
  enhanced, breakable, width=\linewidth,
  colback=gray!5, colframe=gray!50,
  title={Problem, ground truth: $[66, 66, 66]$},
  fonttitle=\bfseries\sffamily, coltitle=black
]
Apply a function to the final input list to generate the output
list. Use any preceding inputs and outputs as examples.

\smallskip
\textbf{Examples:}\\
$[4, 34, 17, 7, 5] \to [17, 7, 5]$\\
$[2, 42, 1, 7, 4, 4, 59, 3, 28, 72] \to [7, 4, 4, 3, 28]$\\
$[88, 8, 55, 12, 4, 2, 75] \to [55, 12, 75]$\\
$[18, 40, 35, 57, 18, 18, 33, 35, 35] \to [35, 33, 35, 35]$

\smallskip
\textbf{Test:} $[46, 94, 66, 98, 66, 66] \to ?$
\end{tcolorbox}

\smallskip
\begin{tcolorbox}[
  enhanced, breakable, width=\linewidth,
  colback=red!3, colframe=red!40,
  title={\textbf{Base}, broad hypothesis exploration without verification},
  fonttitle=\sffamily, coltitle=black
]
\begin{Verbatim}[fontsize=\footnotesize, breaklines, breakanywhere]

[Tries dozens of hypotheses: local maxima removal, suffix extraction, runs of duplicates, threshold filtering...]

Perhaps the function is to **remove all elements that are strictly greater than the previous element?**

Test example 1: [4,34,17,7,5]
- 4->34: 34>4 -> remove. 17<34->keep. 7<17->keep. 5<7->keep.
So output: [17,7,5] — matches!

Check example 2: [...]
Kept: 2, 1, 4, 4, 3 — **not [7,4,4,3,28]** -> fails.

[Tries ~15 more hypotheses, none verified against all examples.]

[...] I think I have to give up and guess [...] based on the only run of duplicates at the end.

```
[66, 66]
```
\end{Verbatim}
\end{tcolorbox}

\smallskip
\begin{tcolorbox}[
  enhanced, breakable, width=\linewidth,
  colback=orange!3, colframe=orange!50,
  title={\textbf{$RL_{individual}$}, contradictions noted but not used},
  fonttitle=\sffamily, coltitle=black
]
\begin{Verbatim}[fontsize=\footnotesize, breaklines, breakanywhere]

[Follows nearly identical exploration. Tests same hypotheses.]

[...] In example 1: no duplicates -> **output should be empty? But output is [17,7,5]** — so not that.

**<- Notes the contradiction explicitly, but continues with the falsified rule anyway.**

[...] I think the function is to keep the last run of identical values of length at least 2. [...] For the test problem: only the last two 66s.

```
[66, 66]
```
\end{Verbatim}
\end{tcolorbox}

\smallskip
\begin{tcolorbox}[
  enhanced, breakable, width=\linewidth,
  colback=green!5, colframe=green!75!black,
  title={\textbf{RACES}, index tracking and rule preservation across all demos},
  fonttitle=\sffamily, coltitle=black
]
\begin{Verbatim}[fontsize=\footnotesize, breaklines, breakanywhere]
[Tests hypotheses systematically, verifying each against ALL examples before proceeding.]

[...] Example 1: from index 2 -> [17,7,5]. Example 2: last 5 elements... **not matching**. [Rejects suffix hypothesis.]

[Shifts to tracking indices explicitly:]

Example 4: output [35,33,35,35] -> indices 2,6,7,8
- 35 appears at indices 2, 7, 8 — appears 3 times total
- 33 appears at index 6 — once

[...] outputs consistently included values appearing multiple times in the input (66 appears at indices 2, 4, 5) -> all three occurrences of the repeated value.

```
[66, 66, 66]
```
\end{Verbatim}
\end{tcolorbox}

\subsection{Enigmata: Sum Skyscraper Logic Puzzle}
\label{app:case-skyscraper}

\textbf{Pattern illustrated.} Stateful constraint propagation with
deliberate decomposition. $RL_{RACES}$ switches the search axis
(rows $\to$ columns) so that the smaller search space is exhausted
first, then propagates the resulting fixed values to constrain the
remaining cells; both Base and $RL_{individual}$ commit to an
incomplete row-wise search and incorrectly conclude that no solution
exists.

\smallskip
\begin{tcolorbox}[
  enhanced, breakable, width=\linewidth,
  colback=gray!5, colframe=gray!50,
  title={Problem, Sum Skyscraper $4{\times}4$},
  fonttitle=\bfseries\sffamily, coltitle=black
]
Fill a $4\times 4$ grid with heights $1$--$4$ so that each row and
column is a permutation of $\{1,2,3,4\}$ and the visibility-sum
clues on all four sides are satisfied (a clue gives the total height
of buildings visible from that side; taller buildings block shorter
ones behind them).

\smallskip
\textbf{Clues} (top, left, right, bottom):
\begin{verbatim}
7 9 4 5
7 9 4 5
5 4 7 9
5 4 7 9
\end{verbatim}

\smallskip
\textbf{Ground-truth answer:}
$\bigl[\,[3,2,4,1],\,[2,3,1,4],\,[4,1,2,3],\,[1,4,3,2]\,\bigr]$.
\end{tcolorbox}

\smallskip
\begin{tcolorbox}[
  enhanced, breakable, width=\linewidth,
  colback=red!3, colframe=red!40,
  title={\textbf{Base}, visibility assumption error in row-wise casework},
  fonttitle=\sffamily, coltitle=black
]
\begin{Verbatim}[fontsize=\footnotesize, breaklines, breakanywhere]
[Correctly identifies visibility-sum semantics. Attempts row-by-row search for Row 1 (left=7, right=5).]

[...tries ~15 permutations...]

Try [3,2,4,1]:
- Left: 3->visible, 2<3->blocked, 4>3->visible -> sum=3+4=7 $\checkmark$
- Right: 1->visible, 4>1->visible -> sum=1+4=5 $\checkmark$

**<- Actually finds a valid Row 1!** But then analyzes right-sum=5 in general:

Case 2: d=2, c=3: row [a,b,3,2], need left=7.
Available {a,b}={1,4}.

Option 1: a=1, b=4 -> left: 1+4=5  x 
Option 2: a=4, b=1 -> left: 4 only  x 

<- KEY ERROR: Ignores [3,2,4,1] found moments earlier. Concludes "no valid solution."
\end{Verbatim}
\end{tcolorbox}

\smallskip
\begin{tcolorbox}[
  enhanced, breakable, width=\linewidth,
  colback=orange!3, colframe=orange!50,
  title={\textbf{$RL_{individual}$}, incomplete row enumeration},
  fonttitle=\sffamily, coltitle=black
]

\begin{Verbatim}[fontsize=\footnotesize, breaklines, breakanywhere]

[Also searches row-by-row. Tests many permutations for Row 1 but never tests [3,2,4,1].]

[...] Try [3,1,4,2]: Left=7 $\checkmark$, Right=6  x 
Try [3,4,1,2]: Left=7 $\checkmark$, Right=6  x 

[Moves to Row 4 (right=9):] "No sequence gives sum = 9."

<- Fails to test [4,3,1,2]: right-to-left visibility: 2->3>2->4>3 -> sum=2+3+4=9. Concludes "No valid solution."

\end{Verbatim}
\end{tcolorbox}

\smallskip
\begin{tcolorbox}[
  enhanced, breakable, width=\linewidth,
  colback=green!3, colframe=green!75!black,
  title={\textbf{RACES}, column-wise decomposition with cascading propagation},
  fonttitle=\sffamily, coltitle=black
]
\begin{Verbatim}[fontsize=\footnotesize, breaklines, breakanywhere]

[**Key strategic choice:** decomposes column-by-column instead of row-by-row.]

**Column 1 (top=7, bottom=5):** Enumerates all 24 permutations systematically.

[...] (3,2,4,1): Top: 3+4=7 $\checkmark$. Bottom: 1->visible, 4>1->visible -> sum=1+4=5 $\checkmark$

**-> Unique valid assignment: A=3, E=2, I=4, M=1**

**Column 2 (top=9, bottom=4):** Bottom=4 forces N=4 (only tallest visible). Then:
- (2,1,3,4): top sum=2+3+4=9 $\checkmark$
- (2,3,1,4): top sum=2+3+4=9 $\checkmark$

**Column 3 (top=4):** Forces C=4 (only way to get sum=4). Then bottom=7 yields two options for (G,K,O).

**Column 4 (bottom=9):** Finds (1,4,3,2) and (4,3,1,2) both valid.

[**Constraint propagation:**]
Row 1: A=3, B=2, C=4 -> D must be **1** (only value left).
-> Fixes Column 4 to Option X: D=1, H=4, L=3, P=2.

Row 2: E=2, H=4 -> F,G $\in$ {1,3}. But E=2 already in row -> **G$\neq$2** -> G=1 -> F=3.
-> Fixes Column 3 Option A and Column 2 Option 2.

[**Full verification of all 16 clues** — all pass.]

```
3 2 4 1
2 3 1 4
4 1 2 3
1 4 3 2
```
\end{Verbatim}
\end{tcolorbox}


\newpage

\end{document}